# How Much Temporal Modeling is Enough? A Systematic Study of Hybrid CNN-RNN Architectures for Multi-Label ECG Classification

**Alireza Jafari [a, \*], Fatemeh Jafari [b]**

[a] Department of Biomedical Engineering, South Tehran Branch, Islamic Azad University, Tehran, Iran

st_ar_jafari@azad.ac.ir

[b] Department of Computer Engineering, Faculty of Engineering and Technology, Islamic Azad University, Gorgan Branch, Gorgan, Iran

f.jafari9254@iau.ir

**Abstract**

Accurate multi-label classification of electrocardiogram (ECG) signals remains challenging due to the coexistence of multiple cardiac conditions, pronounced class imbalance, and long-range temporal dependencies in multi-lead recordings. Although recent studies increasingly rely on deep and stacked recurrent architectures, the necessity and clinical justification of such architectural complexity have not been rigorously examined. In this work, we perform a systematic comparative evaluation of convolutional neural networks (CNNs) combined with multiple recurrent configurations, including LSTM, GRU, Bidirectional LSTM (BiLSTM), and their stacked variants, for multi-label ECG classification on the PTB-XL dataset comprising 23 diagnostic categories. The CNN component serves as a morphology-driven baseline, while recurrent layers are progressively integrated to assess their contribution to temporal modeling and generalization performance. Experimental results indicate that a CNN integrated with a single BiLSTM layer achieves the most favorable trade-off between predictive performance and model complexity. This configuration attains superior Hamming loss (0.0338), macro-AUPRC (0.4715), micro-F1 score (0.6979), and subset accuracy (0.5723) compared with deeper recurrent combinations. Although stacked recurrent models occasionally improve recall for specific rare classes, our results provide empirical evidence that increasing recurrent depth yields diminishing returns and may degrade generalization due to reduced precision and overfitting. These findings suggest that architectural alignment with the intrinsic temporal structure of ECG signals, rather than increased recurrent depth, is a key determinant of robust performance and clinically relevant deployment.

## 1. Introduction

The electrocardiogram (ECG) is the cornerstone of cardiac diagnosis, yet its clinical utility is often limited by interpretive subjectivity and reliance on specialized expertise. Despite recent advances in artificial intelligence for ECG analysis, many deep learning models operate as opaque systems, lacking transparent links to established clinical reasoning. This absence of interpretability critically impedes physician trust and practical deployment [1]. The proliferation of artificial intelligence (AI), the Internet of Things (IoT), and advanced mobile systems is fundamentally restructuring medical care delivery. Contemporary innovations, including wearable devices and mobile-enabled sensor networks, are catalyzing a transition from institution-based healthcare frameworks to models that are intrinsically centered on the individual patient [2, 3]. This technological integration facilitates enhanced early detection capabilities, improved healthcare access, and a reduction in associated costs. Cardiovascular diseases (CVDs), which represent the predominant cause of global mortality, constitute a significant public health challenge in the modern era [4]. The global management of CVDs—including coronary artery disease, heart failure, arrhythmias, and stroke—imposes a substantial economic strain on healthcare infrastructures worldwide [5, 6]. The electrocardiogram (ECG) provides a non-invasive, rapid, and reliable modality for assessing cardiac electrophysiology, establishing its utility in the identification of various CVDs. However, the interpretation of conventional ECG tracings is contingent upon specialist expertise, a process that remains predominantly manual. This reliance introduces delays, elevates costs, and creates significant barriers to care in regions with limited access to cardiologists [7]. The emergence of IoT-facilitated ECG apparatus, integrated with mobile and wireless connectivity and edge computing processors, presents a scalable paradigm for decentralized cardiac surveillance. These systems enable remote physiological monitoring within domestic environments, demonstrating particular promise for deployment in resource-constrained settings [8, 9]. Recent advances in deep learning have transformed the analysis of complex biomedical signals. Deep learning excels in interpreting electrocardiogram (ECG) signals and their temporal dynamics, with federated learning enabling privacy-preserving applications such as heart sound anomaly detection [10]. In self-supervised representation learning for label-scarce ECGs, data augmentation—through transformations like amplitude scaling and temporal permutation—promotes invariant, robust features that generalize across diverse patient cohorts and recording conditions [11]. Concurrently, the SegSE block introduces spatially adaptive recalibration for fully convolutional networks in semantic segmentation, overcoming global channel-wise attention limitations (e.g., Squeeze-and-Excitation) via dilated convolutions and local modulation to enhance feature discriminability in lesion analysis [12]. Notable advancements have been achieved through transfer learning, where architectures including VGG16, ResNet, Inception, and

EfficientNet have been leveraged as feature extractors from ECG spectrograms and other high-dimensional representations [13]. Recent studies [14, 15] have underscored the clinical importance of automated arrhythmia detection from ECG signals, demonstrating that deep learning–based methods can overcome the inherent limitations of traditional approaches relying on hand-crafted features. However, effectively modeling long-term temporal dependencies in multi-lead ECG recordings remains an open challenge. The analysis of temporal patterns in electrocardiogram (ECG) waveforms inherently benefits from sequence modeling approaches. Among these, recurrent neural network (RNN) derivatives, specifically Long Short-Term Memory (LSTM) and Bidirectional LSTM (BiLSTM) architectures, are employed to capture long-range dependencies within physiological time-series data. Empirical benchmarking on large-scale public datasets provides critical validation for these models. The comprehensive benchmark established by Strodthoff et al. [16] on the PTB-XL dataset demonstrated that both LSTM and BiLSTM models deliver competitive diagnostic performance across multiple ECG classification tasks, with the bidirectional variant often achieving a marginal performance advantage. Despite the widespread use of hybrid CNN–RNN architectures in ECG analysis, there is currently limited systematic evidence on whether increasing recurrent depth or combining heterogeneous recurrent units consistently improves multi-label classification performance on large-scale clinical datasets. Most prior studies implicitly assume that deeper or more complex recurrent structures lead to superior temporal modeling and improved diagnostic accuracy. However, this assumption has not been rigorously validated under severe class imbalance and true multi-label settings. The robustness and reproducibility of these findings were subsequently supported by an independent replication study [17]. While Gated Recurrent Unit (GRU) networks were also evaluated in the initial benchmark [16], their detailed comparative results were less extensively reported. Importantly, much of the existing literature implicitly assumes that increasing architectural complexity—such as stacking multiple recurrent layers or combining different recurrent units—leads to superior diagnostic performance. However, this assumption has not been rigorously validated in large-scale, clinically annotated datasets, particularly under multi-label conditions where multiple abnormalities may coexist and severe class imbalance is common. Increasing architectural complexity can introduce practical challenges for clinical deployment, such as higher computational requirements, reduced model transparency, and greater susceptibility to overfitting. In this work, we address these limitations by adopting a convolutional neural network (CNN) as a principled baseline for learning morphology-driven ECG features and systematically extending this baseline with different recurrent architectures to model temporal dependencies. Specifically, we evaluate combinations of CNN with LSTM, BiLSTM, and GRU layers, as well as stacked recurrent configurations, under a unified experimental protocol. This design enables a controlled assessment of how incremental temporal modeling capacity contributes to classification performance across both frequent and rare diagnostic categories. Furthermore, we incorporate data augmentation to improve robustness under limited and imbalanced training data and employ an attention-based feature reweighting mechanism to adaptively emphasize diagnostically salient latent representations. Together, these components allow the proposed framework to balance morphological feature extraction, temporal context modeling, and adaptive feature selection without assuming that increased architectural complexity inherently yields superior performance. Using the PTB-XL dataset with 23 diagnostic classes, we provide a comprehensive comparative analysis of these architectures in a multi-label setting. This systematic evaluation offers empirical evidence regarding the trade-offs between representational capacity, generalization performance, and architectural complexity, thereby informing the design of practical and clinically deployable deep learning models for automated ECG interpretation. Accordingly, the central research question of this study is: Does increasing recurrent depth and architectural complexity provide systematic and clinically meaningful performance gains for multi-label ECG classification, or can simpler controlled hybrid architectures achieve comparable or superior generalization?

This paper is organized as follows: Section 2 describes the proposed methodology, Section 3 presents the experimental results, Section 4 discusses the key findings, and Section 5 concludes the paper.

## 2. Materials and Methods

### 2.1. Dataset

All experiments were conducted using the PTB-XL dataset (version 1.0.3), following the general data preparation principles proposed by Wagner et al. [18]. After excluding recordings associated with invalid or non-diagnostic SCP codes, the final dataset comprised 21,799 high-quality 12-lead electrocardiogram (ECG) recordings. Each record represents a 10-second acquisition sampled at 100 Hz using the low-resolution signals provided in the original release. To ensure an unbiased and clinically realistic evaluation, all data partitioning was performed on a patient-wise basis, thereby preventing any overlap between training, validation, and test cohorts.

### 2.2. Diagnostic Subclasses and Multi-Label Formulation

A total of 23 diagnostic subclasses were derived from the SCP codes provided in the PTB-XL annotations. Each ECG recording was associated with one or more diagnostic subclasses, resulting in a multi-label classification setting. The resulting class distribution reflects real-world clinical prevalence, with the most frequent class (NORM) comprising 9,514 recordings and the rarest class (PMI) comprising 17 recordings, corresponding to an approximate imbalance ratio of 560:1. Overall, 28.6% of the recordings were assigned multiple diagnostic labels, with an average of 1.39 labels per sample.

### 2.3. Data Splitting and Class Imbalance

The official 10-fold stratified splits provided with the PTB-XL dataset were adopted to ensure reproducibility and strict patient separation. Folds 1–8 were used for model training (17,418 recordings), fold 9 for validation (2,183 recordings), and fold 10 for independent testing (2,198 recordings). This split strategy preserves the multi-label class distribution across all subsets while guaranteeing that no patient appears in more than one partition [18]. Fig. 1 illustrates the inherent class distribution imbalance within the PTB-XL dataset, ranging from highly prevalent normal rhythms to rare diagnostic subclasses with fewer than 25 samples. This long-tail distribution represents a primary challenge for training unbiased and generalizable deep learning models. Consequently, model design and evaluation were conducted with explicit consideration of imbalance-related effects on threshold-dependent performance metrics.

the corresponding standard deviation, resulting in zero-centered signals with comparable dynamic ranges across leads. This procedure prevents information leakage while preserving inter-lead relationships. Apart from this standardization step, no additional preprocessing or resampling was applied, and the signals were otherwise used as provided in the original PTB-XL release. Signals are stored in single-precision floating-point format to reduce memory consumption and facilitate GPU acceleration. No handcrafted features are extracted; instead, the model directly learns hierarchical representations from raw ECG signals.

### 2.5. Data Augmentation

To improve robustness and reduce overfitting, simple time-domain augmentation [19] was applied during

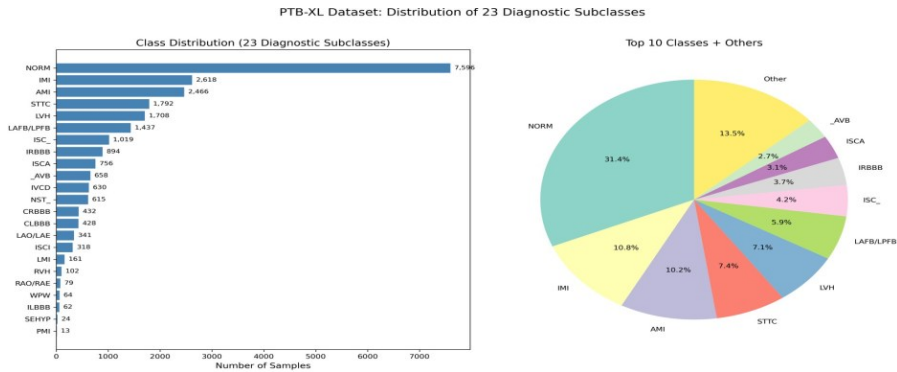

**Fig. 1.** Distribution of Diagnostic Subclasses in the PTB-XL Dataset

### 2.4. Signal Normalization and Representation

Each ECG recording is represented as a two-dimensional tensor $\mathbf{X} \in \mathbf{R}^{T \times 12}$, where T denotes the temporal length of the signal and 12 corresponds to the number of ECG leads. All recordings were independently standardized on a per-lead basis using statistics computed exclusively from the training subset. For each lead, the mean value was subtracted and the signal was scaled by training. The augmentation strategy includes additive Gaussian noise to simulate acquisition-related perturbations and random amplitude scaling to model inter-subject variability. Augmentation was applied stochastically at the batch level and was disabled during validation and testing to ensure unbiased evaluation. These transformations were restricted to physiologically plausible ranges to preserve the clinical integrity of ECG waveforms.

## 2.6. Memory-Efficient Data Loading

Due to the large size of the dataset, ECG signals were loaded using a custom batch generator implemented via the Keras Sequence interface. This approach enables efficient batch-wise data loading, supports shuffling at the epoch level, and ensures stable memory usage during training.

**TABLE 1**

MODEL ARCHITECTURE CONFIGURATION.

| Layer Type | Output Shape | Parameters | Notes |
| --- | --- | --- | --- |
| Input | (1000, 12) | 0 | ECG 12-lead waveform |
| Conv1D-64 | (1000, 64) | 11,264 | Kernel=15, ReLU, BatchNorm |
| MaxPooling1D | (500, 64) | 0 | Pool size=2 |
| SpatialDropout1D | (500, 64) | 0 | Rate=0.1 |
| Conv1D-128 | (500, 128) | 82,048 | Kernel=10, ReLU, BatchNorm |
| MaxPooling1D | (250, 128) | 0 | Pool size=2 |
| SpatialDropout1D | (250, 128) | 0 | Rate=0.1 |
| Conv1D-256 | (250, 256) | 164,096 | Kernel=5, ReLU, BatchNorm |
| MaxPooling1D | (125, 256) | 0 | Pool size=2 |
| LSTM | (125, 128) | 197,120 | Temporal modeling |
| BiLSTM | (125, 256) | 263,168 | Bidirectional temporal modeling |
| GRU | (125, 128) | 148,992 | Efficient temporal modeling |
| GlobalAveragePooling1D | (256) | 0 | Temporal summary |
| Dense-512 | (512) | 131,584 | ReLU, BatchNorm, Dropout=0.5 |
| Dense-256 | (256) | 131,328 | ReLU, BatchNorm, Dropout=0.5 |

## 2.7. Model Architecture Overview

The proposed framework integrates convolutional and recurrent neural network components to jointly capture morphological and temporal characteristics of ECG signals. The architecture consists of four main stages: (1) convolutional feature extraction, (2) sequential temporal modeling, (3) global feature aggregation with attention-based reweighting, and (4) fully connected classification layers. This modular design facilitates a systematic evaluation of architectural components while maintaining architectural transparency. The detailed specifications of each stage are provided in Table 1.

## 2.8. Convolutional Feature Extraction (CNN Baseline)

A convolutional neural network (CNN) serves as the baseline architecture and morphological feature extractor. The convolutional backbone comprises three one-dimensional convolutional blocks with increasing filter depths. Each block includes a 1D convolutional layer with He initialization [20], followed by batch normalization, ReLU activation, max-pooling for temporal downsampling, and spatial dropout for regularization. These layers progressively extract local morphological patterns, such as QRS complexes and ST-T characteristics, while reducing temporal redundancy and suppressing noise. This CNN-only configuration establishes a morphology-focused reference model against which the added value of sequential modeling is assessed.

## 2.9. Sequential Temporal Modeling and Systematic Architecture Variants

Temporal dependencies are modeled by integrating recurrent neural network layers with the convolutional feature extractor. To systematically analyze the impact of temporal modeling depth and directionality, multiple hybrid architectures were evaluated. These include CNN–LSTM [21], CNN–BiLSTM [21], CNN–BiLSTM–LSTM, CNN–GRU [22], and CNN–GRU–BiLSTM–

LSTM. In these hybrid configurations, recurrent layers are applied sequentially to the convolutional feature maps to capture temporal dependencies at varying scales. Unidirectional LSTM and GRU layers model long-range temporal structure and rhythm-level dynamics, while bidirectional LSTM layers incorporate contextual information from both past and future time steps within each ECG segment. The inclusion of multiple recurrent configurations enables a controlled and systematic comparison of temporal modeling strategies, allowing assessment of whether additional recurrent depth and bidirectionality provide consistent performance benefits across heterogeneous diagnostic categories. Batch normalization and spatial dropout are consistently applied between recurrent layers to improve training stability and reduce overfitting.

### 2.10. Global Pooling and Attention-Based Feature Reweighting

After temporal modeling, a global average pooling layer aggregates sequence-level features into a fixed-length representation. An attention-based feature reweighting mechanism is subsequently applied. This module consists of two fully connected layers with ReLU and sigmoid activations, producing a feature-wise attention vector that modulates the pooled representation through element-wise multiplication. This adaptive reweighting allows the model to emphasize diagnostically informative latent dimensions while suppressing less relevant components, thereby improving robustness and contextual sensitivity without discarding feature continuity. To explicitly quantify the contribution of the feature-wise gating mechanism, we include an ablation experiment in which the gating module is removed while keeping all other architectural components unchanged.

### 2.11. Classification Head

The classification head consists of two fully connected layers with batch normalization and dropout for regularization. The final output layer uses a sigmoid activation function to independently estimate the probability of each diagnostic subclass, making the model suitable for multi-label classification under severe class imbalance.

### 2.12. Training Configuration

The model is trained using the Adam optimizer with an initial learning rate of 0.001. Binary cross-entropy loss with label smoothing is employed to account for annotation noise and label uncertainty. Mixed-precision training is enabled to accelerate training and reduce GPU memory usage. Early stopping based on validation AUROC is used to prevent overfitting, and the learning rate is adaptively reduced when validation loss plateaus.

### 2.13. Cross-Validation Protocol

Five-fold cross-validation is applied exclusively within the combined training and validation folds (folds 1–9) for model selection and robustness analysis. Final performance is reported only on the official independent test fold (fold 10), which is never used during cross-validation or hyperparameter tuning. In each fold, the model is trained from scratch using identical hyperparameters, and performance metrics are computed on the held-out validation fold. Aggregate statistics across folds are reported. After cross-validation, a final model is trained using the original training set and evaluated on the independent test set.

**TABLE 2**
PERFORMANCE OF SIX DEEP LEARNING ARCHITECTURES—INCLUDING CNN, RECURRENT, AND HYBRID MODELS—ON THE MULTI-LABEL ECG CLASSIFICATION TASK ACROSS 23 DIAGNOSTIC CLASSES, EVALUATED USING MACRO-, MICRO-, AND SUBSET-LEVEL METRICS.

| Model | Hamming Loss ↓ | Macro AUROC ↑ | Macro AUPRC ↑ | Macro F1 ↑ | Macro Precision ↑ | Macro Recall ↑ | Macro Balanced Accuracy ↑ | Micro F1 ↑ | Micro Precision ↑ | Micro Recall ↑ | Subset Accuracy ↑ |
|---|---|---|---|---|---|---|---|---|---|---|---|
| GRU+BiLSTM+LSTM | 0.0354 | 0.9080 | 0.4409 | 0.3496 | 0.4766 | 0.3182 | 0.6509 | 0.6754 | 0.7498 | 0.6144 | 0.5487 |
| GRU | 0.0348 | 0.9195 | 0.4598 | 0.3755 | 0.4611 | 0.3435 | 0.6632 | 0.6864 | 0.7466 | 0.6351 | 0.5537 |
| LSTM+BiLSTM | 0.0344 | 0.9166 | 0.4711 | 0.3998 | 0.5419 | 0.3618 | 0.6723 | 0.6913 | 0.7491 | 0.6417 | 0.5605 |
| BiLSTM | 0.0338 | 0.9202 | 0.4715 | 0.3908 | 0.5505 | 0.3496 | 0.6663 | 0.6979 | 0.7535 | 0.6500 | 0.5723 |
| LSTM | 0.0347 | 0.9231 | 0.4571 | 0.3698 | 0.5392 | 0.3310 | 0.6567 | 0.6886 | 0.7473 | 0.6384 | 0.5605 |
| CNN | 0.0340 | 0.9204 | 0.4721 | 0.3841 | 0.5630 | 0.3439 | 0.6635 | 0.6944 | 0.7543 | 0.6434 | 0.5664 |

## 2.14. Evaluation Metrics

Performance is evaluated using metrics appropriate for multi-label classification, including macro-averaged AUROC, Hamming loss, macro, micro, and weighted F1-scores, subset accuracy, and overall label-wise accuracy. This comprehensive evaluation protocol accounts for both threshold-dependent and threshold-independent performance under severe class imbalance. Unless otherwise stated, a fixed decision threshold of 0.5 is used for all classes to compute threshold-dependent metrics (F1-score, subset accuracy). Additional experiments using class-wise optimized thresholds are reported in the supplementary material to assess the sensitivity of results to threshold selection.

## 2.15. Implementation Details

All experiments were conducted using Python 3 and TensorFlow (version 2.19). Random seeds were fixed to ensure reproducibility. Model training and evaluation were performed on a workstation equipped with an NVIDIA GeForce RTX 3050 GPU (6 GB), 16 GB RAM, and a 4-core CPU. GPU dynamic memory growth was enabled to allow efficient memory utilization.

## 3. Results

Table 2 summarizes the performance of all six deep learning architectures on the multi-label ECG classification task across 23 diagnostic classes using macro-, micro-, and subset-level metrics. The evaluated models include a CNN baseline and multiple recurrent and hybrid configurations, enabling a systematic assessment of how increasing temporal modeling complexity affects overall classification performance. Across all architectures, strong discriminative capability is observed, with macro-averaged AUROC values consistently exceeding 0.90, indicating robust separation between positive and negative samples across diagnostic categories. Among the evaluated models, the BiLSTM and CNN architectures achieve the most balanced overall performance across evaluation criteria. In particular, the BiLSTM model attains the lowest Hamming Loss (0.0338) and the highest Subset Accuracy (0.5723), reflecting fewer label-wise errors and improved joint label assignment in the multi-label setting. It also demonstrates competitive macro- and micro-averaged performance, including Macro AUROC (0.9202), Macro AUPRC (0.4715), and Micro F1 (0.6979), supporting its effectiveness in capturing temporal dependencies relevant to ECG interpretation. The CNN baseline exhibits strong precision-oriented behavior, achieving the highest Macro Precision (0.5630) and Macro AUPRC (0.4721). This highlights the effectiveness of convolutional feature extraction in capturing local morphological characteristics and accurately identifying dominant diagnostic patterns, particularly for majority classes. The competitive performance of the CNN further confirms that local waveform morphology provides a strong foundation for multi-label ECG classification. Hybrid recurrent architectures, including LSTM+BiLSTM and GRU+BiLSTM+LSTM, do not consistently outperform simpler configurations. While these models provide increased representational capacity, their Macro F1 and Subset Accuracy are comparable to, and in some cases lower than, those of the BiLSTM and CNN models. This pattern suggests diminishing returns from increased recurrent depth and indicates that additional architectural complexity does not necessarily translate into systematic performance gains for this dataset. These findings support the use of simpler, well-regularized architectures when balancing performance and model complexity. Overall, no single architecture dominates across all metrics. Instead, the results reveal a trade-off between contextual temporal modeling and local morphological discrimination. BiLSTM-based models tend to improve joint multi-label assignment and micro-averaged recall through enhanced sequence modeling, whereas CNN-based models favor precision and AUPRC through strong local feature discrimination. This systematic comparison validates the architectural design strategy and reduces the likelihood of overfitting-driven model selection.

### 3.1. Class-wise Performance

Fig. 2 presents the normalized class-wise confusion summary. High specificity is observed across most diagnostic categories, with true negative rates exceeding 0.45 for all classes, indicating reliable rejection of non-relevant labels. The NORM class exhibits a relatively higher false positive rate, reflecting occasional misclassification of pathological recordings as normal, a behavior commonly observed in multi-label ECG settings where normal patterns overlap with mild abnormalities. Performance varies substantially across diagnostic subclasses, largely reflecting differences in class prevalence. Minority classes, including PMI, RAO/RAE, and WPW, demonstrate near-perfect precision but variable and often low recall, consistent with their limited representation in the dataset.

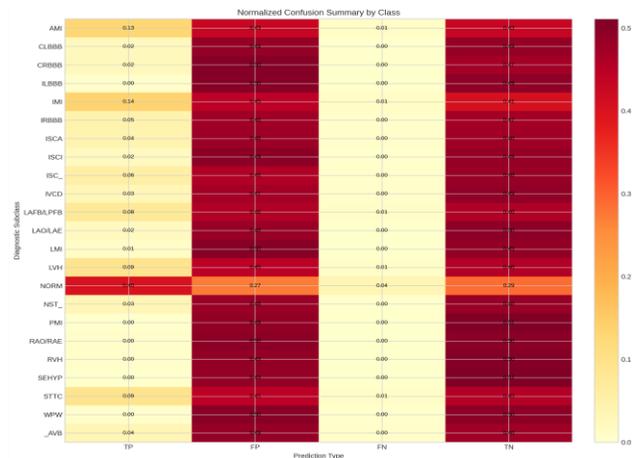

**Fig. 2.** Normalized Confusion Summary by Class

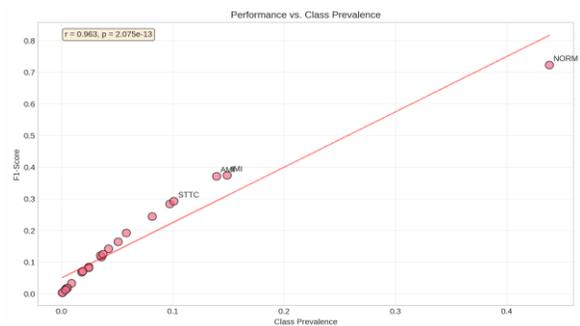

**Fig. 3.** Scatter Plot of Performance versus Class Prevalence

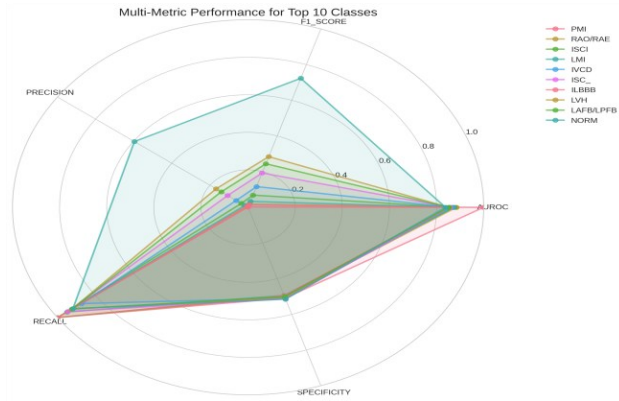

**Fig. 4.** Radar Chart of Multi-Metric Performance for Top 10 Classes

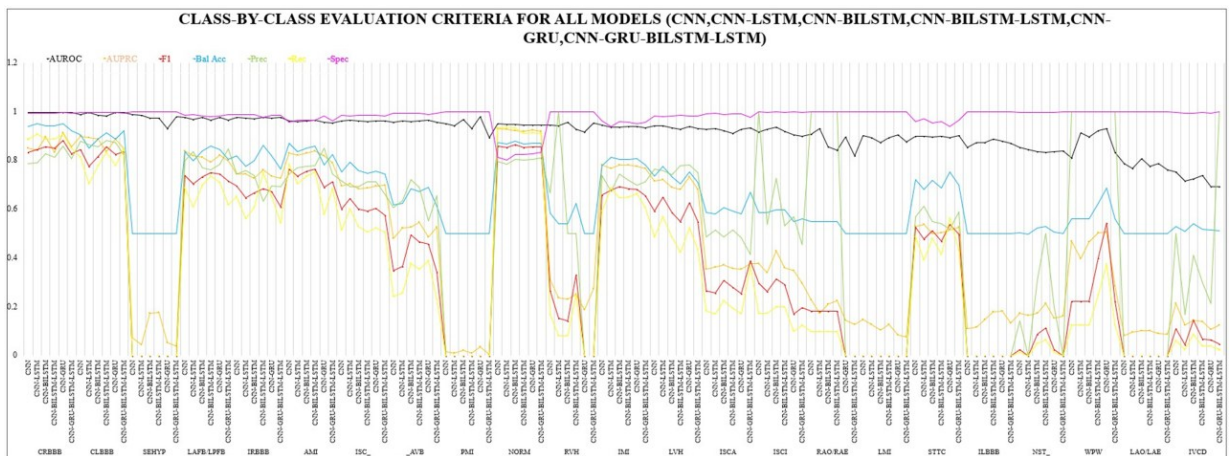

**Fig. 5.** Class-wise Evaluation Metrics for Hybrid CNN-RNN Models

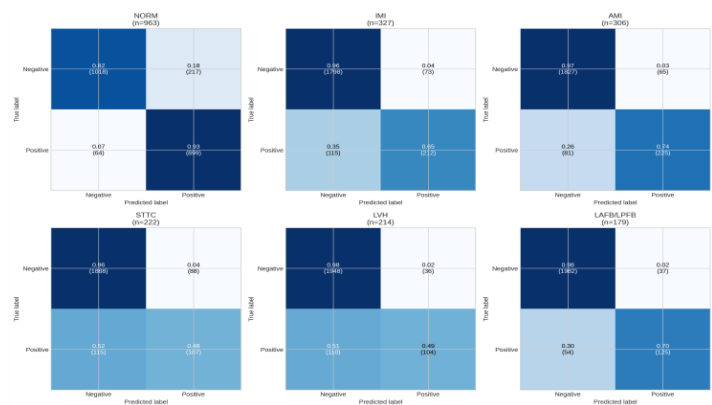

**Fig. 6.** Confusion Matrices for the Top Six Classes by Sample Size

**Fig. 7.** Distribution of Key Performance Metrics Across All Classes

**Fig. 8.** Per-Class AUROC and AUPRC Values, Sorted by AUROC

**Fig. 9.** Precision-Recall Curves for All Diagnostic Classes

**Fig. 10.** Receiver Operating Characteristic Curves for All Diagnostic Classes

This behavior reflects threshold-dependent metric instability under extreme class imbalance rather than poor ranking performance. Fig. 3 illustrates the relationship between class prevalence and F1-score, revealing a strong positive correlation (Spearman $\rho = 0.963$, $p < 0.001$). This confirms that classification performance is strongly influenced by the number of available training samples and highlights the inherent challenges of learning reliable decision boundaries for rare conditions in multi-label medical datasets. Fig. 4–6 provide multi-metric visualizations and confusion matrices for the most prevalent and top-performing classes. These results demonstrate well-balanced precision, recall, specificity, and F1-scores for high-prevalence conditions such as NORM, CRBBB, and CLBBB. In contrast, underrepresented classes exhibit greater variability across metrics, reflecting limited sample support and increased uncertainty in threshold-based predictions. Fig. 7–10 further summarize class-level performance across all diagnostic categories. Consistently high AUROC values are observed for most classes (mean AUROC = 0.920), indicating strong overall discriminative ability. However, metrics sensitive to class prevalence, including F1-score and AUPRC, are notably lower for rare conditions. For example, AUROC exceeds 0.95 for CRBBB, AMI, and CLBBB, whereas low-prevalence classes such as IVCD and LAO/LAE achieve AUROC values of 0.725 and 0.807, respectively. These trends underscore the model's ability to rank positive samples effectively while highlighting the difficulty of achieving high positive predictive value for extremely rare diagnoses.

### 3.2. Key Observations

First, classification performance is strongly dependent on class prevalence. High-prevalence classes consistently achieve superior F1, recall, and precision, while rare classes exhibit reduced threshold-dependent metrics despite often maintaining high AUROC values. Second, increasing architectural complexity through stacked or hybrid recurrent configurations does not guarantee improved performance. In several cases, simpler BiLSTM or CNN models achieve comparable or superior robustness, indicating diminishing returns from deeper recurrent stacking. Third, a systematic trade-off is observed between temporal context modeling and local feature discrimination. CNN-based models favor precision and AUPRC through strong morphological feature extraction, whereas BiLSTM-based models improve subset accuracy and micro-averaged recall by better capturing sequential context. Finally, the observed disparity between AUROC and F1/AUPRC values reflects the known challenges of multi-label classification under severe class imbalance. This highlights the importance of interpreting threshold-dependent metrics with caution for rare conditions and underscores the practical limitations inherent to clinical datasets with highly skewed label distributions.

### 4. Discussion

Automated ECG interpretation poses a unique challenge due to the coexistence of short-lived morphological events and long-range temporal dependencies, together with the frequent presence of overlapping cardiac abnormalities. To address these characteristics, this study adopts a systematically constructed hybrid architecture that integrates convolutional and recurrent components within a balanced and clinically grounded design framework. Rather than assuming that increased architectural complexity inherently leads to superior performance, multiple recurrent configurations were evaluated to characterize how different levels of temporal modeling contribute to classification outcomes under realistic multi-label and imbalanced conditions. The convolutional component serves as the morphological foundation of the framework by capturing localized waveform characteristics, including QRS morphology, ST-segment deviations, and T-wave variations. These features represent core diagnostic cues that can be effectively modeled through learned convolutional filters. Building on this representation, recurrent layers aggregate temporal information across cardiac cycles, enabling the model to encode rhythm-level and sequence-dependent patterns that extend beyond the receptive field of convolutional kernels. This hierarchical organization reflects a deliberate balance between expressive temporal modeling and parameter efficiency, ensuring that sequential dependencies are captured without introducing unnecessary architectural depth. A distinctive design element is the feature-wise gating mechanism applied after global temporal aggregation. In contrast to time-indexed attention, this module operates along the feature dimension, adaptively recalibrating latent channels based on their learned diagnostic relevance. This formulation allows the network to emphasize informative feature dimensions while suppressing redundant or noisy components, thereby improving robustness and enhancing interpretability at the representation level. By preserving full feature continuity through soft gating, the model avoids brittle hard selection while enabling context-aware feature modulation. The adoption of a multi-label formulation directly reflects clinical ECG interpretation practice, where multiple abnormalities frequently coexist within a single recording. Independent sigmoid-activated outputs preserve meaningful label co-occurrence and prevent information loss associated with mutually exclusive class assumptions. This formulation is particularly important for conduction disorders and ischemic patterns that often overlap. Performance evaluation using complementary metrics, including AUROC, AUPRC, F1-score, and subset accuracy, provides a balanced assessment under severe class imbalance and avoids misleading conclusions based on threshold-dependent measures alone.

### 4.1. Comparison with Prior Work

To contextualize the proposed framework, Table 3 compares it with two representative approaches: the lightweight multi-receptive-field CNN proposed by Feyisa et al. [23] and the MSW-Transformer model introduced by Zhang et al. [24]. In contrast to Feyisa et al., which primarily relies on convolutional feature extraction with implicit temporal modeling, the proposed method explicitly incorporates recurrent mechanisms to capture sequential dependencies. Compared to the transformer-based approach of Zhang et al., which assumes a multi-class formulation and relies on high-capacity global attention, the proposed framework supports true multi-label learning and maintains architectural restraint through controlled hybridization. Importantly, the proposed method integrates feature-wise gating to enable adaptive feature recalibration, a capability not explicitly addressed in either prior work. This design choice aligns temporal context modeling with feature-level diagnostic relevance, providing a balanced alternative to both purely convolutional and high-capacity transformer-based strategies. As a result, the proposed architecture achieves a favorable trade-off between interpretability, temporal modeling fidelity, and robustness under class imbalance.

**TABLE 3**
ANALYTICAL COMPARISON BETWEEN THE PROPOSED METHOD AND RELATED ECG CLASSIFICATION APPROACHES

| Dimension | Feyisa et al. (2022) | Zhang et al. (2023) | Proposed Method |
|---|---|---|---|
| Learning paradigm | Multi-label | Multi-class | Multi-label |
| Label overlap modeling | Explicit | - Not supported | Explicit |
| Temporal dependency modeling | Implicit (CNN) | Global attention | Explicit (BiLSTM + GRU + LSTM) |
| Feature recalibration / interpretability | - | - | Feature-wise gating |
| Architectural philosophy | Lightweight CNN | High-capacity Transformer | Balanced hybrid |
| Model depth sensitivity | Low | High | Controlled (ablation) |
| Class imbalance treatment | Limited | Not addressed | Comprehensive |
| Primary evaluation metrics | AUROC, F1 | Accuracy, AUROC | AUROC, AUPRC, F1, Subset Acc |
| Clinical interpretability | Moderate | Limited | High |

### 4.2. Model Complexity and Generalization

The experimental results demonstrate that increasing recurrent depth or stacking heterogeneous recurrent layers does not consistently yield performance gains. In particular, simpler hybrid configurations, such as CNN–BiLSTM, frequently achieve competitive or superior subset accuracy and Hamming loss compared to deeper composite models. These findings indicate that once sufficient temporal context is captured, additional architectural complexity produces diminishing returns. This observation supports the architectural philosophy of controlled hybrid design, where complexity is introduced only when it provides measurable benefit. The results are consistent with prior observations emphasizing the effectiveness of efficient CNN-based and moderate recurrent architectures, and they contrast with high-capacity models that may be more prone to overfitting in severely imbalanced multi-label settings.

### 4.3. Effect of Class Imbalance

Class prevalence strongly influences threshold-dependent metrics, as reflected in the observed relationship between support size and F1-score or recall. High-prevalence classes, such as NORM and CRBBB, achieve robust precision–recall performance, while rare categories exhibit reduced F1-scores despite maintaining acceptable AUROC values. This divergence highlights the limitation of single-metric evaluation and reinforces the importance of complementary measures such as AUPRC and subset accuracy for fair assessment under extreme imbalance. These results indicate that, for low-support classes, the model retains discriminative capacity at the ranking level while conservative decision thresholds suppress positive predictions. This behavior reflects inherent data limitations rather than a fundamental failure of representation learning.

### 4.4. Clinical Implications

Class-wise performance patterns suggest clinically meaningful behavior. High specificity across most diagnostic categories indicates a conservative decision strategy that limits false alarms. The elevated bias toward normal predictions may reduce unnecessary clinical alerts, although it also underscores the importance of clinician oversight for subtle abnormalities. For several critical conditions, high precision confirms that positive predictions are reliable when produced, supporting the use of the model as a clinical decision-support tool rather than a fully autonomous diagnostic system. The feature-wise gating mechanism further enhances clinical relevance by emphasizing diagnostically salient latent dimensions and suppressing redundant signals. This adaptive representation contributes to robustness in noisy clinical environments and provides a more interpretable internal structure for downstream analysis.

### 4.5. Limitations and Future Directions

A primary limitation of this study is the absence of external validation on independent datasets. Although PTB-XL provides a large and diverse benchmark, generalization across institutions, acquisition protocols, and patient populations cannot be guaranteed. Future work should therefore include multi-center external validation to assess real-world robustness.

### 4.6. Summary

In summary, this study presents a hybrid CNN–recurrent framework with feature-wise gating for multi-label ECG classification under severe class imbalance. By jointly modeling local morphology and temporal dependencies, the proposed architecture achieves robust generalization. Importantly, the results provide empirical evidence that increasing recurrent depth yields diminishing returns and may degrade generalization. These findings support a controlled hybrid design principle, in which architectural complexity is increased only when justified by measurable performance gains, offering a practical and scientifically grounded guideline for real-world ECG modeling.

## 5. Conclusions

This study presented a systematic investigation of hybrid deep learning architectures for multi-label classification of 12-lead ECG signals. By integrating convolutional feature extraction with sequential modeling components, the proposed framework was designed to jointly capture local morphological characteristics and longer-range temporal structures inherent in cardiac electrical activity. Experimental results across a broad set of diagnostic classes demonstrate that hybrid architectures provide improved robustness and discriminative capability compared to standalone convolutional models. Among the evaluated configurations, the CNN–BiLSTM architecture achieved the most favorable balance between predictive performance and architectural complexity, exhibiting stable behavior across both common and less frequent diagnostic categories. These findings indicate that bidirectional temporal context plays an important role in ECG interpretation, particularly for conditions characterized by subtle and context-dependent waveform variations. The analysis further shows that increasing architectural depth does not necessarily result in systematic performance gains. While deeper hybrid configurations were included to enable structured architectural comparison, the results reveal diminishing returns and, in some cases, reduced generalization. This highlights the importance of principled model design rather than excessive stacking of sequential components, especially in clinical signal processing settings where data imbalance and reliability are critical considerations. Despite the strong overall performance, challenges remain for rare diagnostic classes with limited sample support. The constrained precision–recall behavior observed for these categories suggests that future improvements may require data-centric strategies such as targeted augmentation, cost-sensitive optimization, or semi-supervised learning. Overall, the proposed framework provides a scalable experimental reference and a practically viable baseline for multi-label

ECG classification, while highlighting important architectural trade-offs relevant for clinical decision-support system design.

## AUTHORSHIP CONTRIBUTION STATEMENT

Alireza Jafari: Conceptualization, Methodology, Software, Writing – original draft, Writing – review & editing. Fatemeh Jafari: Conceptualization, Methodology.

## DECLARATION STATEMENT

This research received no external funding from academic institutions, public agencies, or commercial entities. The authors explicitly declare no financial or non-financial interests—whether direct or indirect—that could affect the integrity of the study's design, execution, or interpretation. The investigation was carried out autonomously, ensuring intellectual independence and impartiality throughout the process.

**Data and Code Access**

All analytical scripts and computational tools developed for this work are provided in the accompanying supplementary materials and can be accessed via the following repository:

Dataset link: https://physionet.org/content/ptb-xl/1.0.1/

Code link: https://github.com/alireza720/Multi-Label-Classification